\documentclass[letterpaper, 10 pt, conference]{ieeeconf}  

\IEEEoverridecommandlockouts                              

\usepackage[utf8]{inputenc}
\usepackage{graphicx}
\usepackage{amssymb}
\usepackage{amsmath}
\usepackage{xcolor}

\usepackage{booktabs} 
\usepackage{caption} %
\usepackage{algorithm}
\usepackage{algorithmic}
\usepackage{mdframed}
\usepackage{verbatim}
\usepackage[font=footnotesize]{caption}
\definecolor{deepblue}{rgb}{0.1, 0.3, 0.6}   
\definecolor{lightgreen}{HTML}{66CC66}

\title{\LARGE \bf
Automated Hybrid Reward Scheduling via Large Language Models for Robotic Skill Learning
}

\author{Changxin Huang, Junyang Liang, Yanbin Chang, Jingzhao Xu, Jianqiang Li
\thanks{*This work is supported in part by the National Natural Science Funds for Distinguished Young Scholar under Grant 62325307, in part by the National Natural Science Foundation of China under Grants 6240020443, 62073225, 62203134, in part by the Natural Science Foundation of Guangdong Province under Grants 2023B1515120038, in part by Shenzhen Science and Technology Innovation Commission (20231122104038002, 20220809141216003, KJZD20230923113801004), 
in part by the Scientific Instrument Developing Project of Shenzhen University under Grant 2023YQ019 (\textit{Corresponding author: Jianqiang Li})}
\thanks{All authors are with National Engineering Laboratory for Big Data System Computing Technology, Shenzhen University, Shenzhen 518061, China.
        {\tt\small huangchx@szu.edu.cn, \{liangjunyang2018, changyanbin2023\}@email.szu.edu.cn,  xujingzhao2024@gmail.com, lijq@szu.edu.cn}
        }%
}

\begin{document}

\maketitle{}
\thispagestyle{empty}
\pagestyle{empty}

\begin{abstract}

Enabling a high-degree-of-freedom robot to learn specific skills is a challenging task due to the complexity of robotic dynamics. Reinforcement learning (RL) has emerged as a promising solution; however, addressing such problems requires the design of multiple reward functions to account for various constraints in robotic motion. Existing approaches typically sum all reward components indiscriminately to optimize the RL value function and policy. We argue that this uniform inclusion of all reward components in policy optimization is inefficient and limits the robot’s learning performance. To address this, we propose an Automated Hybrid Reward Scheduling (AHRS) framework based on Large Language Models (LLMs). This paradigm dynamically adjusts the learning intensity of each reward component throughout the policy optimization process, enabling robots to acquire skills in a gradual and structured manner. Specifically, we design a multi-branch value network, where each branch corresponds to a distinct reward component. During policy optimization, each branch is assigned a weight that reflects its importance, and these weights are automatically computed based on rules designed by LLMs. The LLM generates a rule set in advance, derived from the task description, and during training, it selects a weight calculation rule from the library based on language prompts that evaluate the performance of each branch. Experimental results demonstrate that the AHRS method achieves an average  6.48\% performance improvement across multiple high-degree-of-freedom robotic tasks.

\end{abstract}


\section{INTRODUCTION}
Embodied intelligent robots learn skills by controlling interactions with their environment\cite{o2024open}, acquiring human-desired skills from interaction data to perform specific actions or tasks~\cite{schaul2015universal, haarnoja2024learning}. However, as the degrees of freedom and dynamic complexity of robots increase, this task becomes more challenging. This complexity necessitates the imposition of additional constraints as optimization goals. Reinforcement learning (RL) optimizes policies by maximizing cumulative rewards \cite{henderson2018deep, oh2020discovering}, effectively transforming each robot constraint into a reward component. This approach has demonstrated success in various robotic tasks, such as legged locomotion \cite{smith2022legged, chen2024reinforcement}, dexterous hand operation \cite{ze2024h, arunachalam2023dexterous}, and robotic manipulation \cite{chen2023bi, tung2021learning}. Nevertheless, current RL methods optimize policies by summing all reward components, compelling robots to learn multiple optimization objectives simultaneously and in parallel, which is difficult for robots and will limit the learning efficiency \cite{mnih2013playing, schulman2017proximal}.

Van et al. propose a Hybrid Reward Architecture (HRA) \cite{van2017hybrid}, which decomposes the reward function and aims to learn a separate value functions, each associated with a distinct reward component. Learning separate value functions in this manner has been shown to facilitate more effective learning. Huang et al. further advanced this approach by introducing a hybrid and dynamic policy gradient (HDPG) method \cite{huang2022reward}, which utilizes dynamic priorities to adjust the contribution of each reward branch during policy optimization. This technique enables robots to prioritize learning components that exhibit rapid reward accumulation, allowing the robot to focus on ``simpler" components before tackling more ``challenging" ones. However, this method heavily depends on human expertise to design the rules for dynamic weight calculation, and the same rules may not guarantee competitive performance across different robotic tasks.

To alleviate the challenge of manually designing dynamic weight rules, this paper proposes an Automated Hybrid Reward Scheduling (AHRS) framework, a language-instructed approach for automatically generating dynamic weight rules.

Similar to HDPG \cite{huang2022reward}, this paper first decomposes the original reward function into several independent reward components (e.g., torque reward, angular velocity reward, linear velocity reward, etc.). A multi-branch value network is then introduced, with each branch dedicated to learning a corresponding reward component. During the training process, we utilize language prompts for large language models (LLMs) to guide the selection of an appropriate rule from a rule buffer, which is then used to calculate the importance of each reward component. These weights are assigned to the respective branches of the multi-branch value network, which subsequently informs the policy training. Two critical issues are addressed in this approach: \textbf{1) How to automate the construction of a dynamic weight rule repository? 2) How should the language prompts for the LLM be designed to optimally select appropriate rules?}

For the first issue, we propose a language-instructed rule generation method. This approach uses the robot task description, environment description, and information about each reward component as prompts to query the large language model (LLM). By leveraging the LLM's powerful reasoning and generative capabilities, a dynamic weight computation rule repository is generated. This repository is constructed prior to the commencement of RL training.

For the second issue, we introduce a policy evaluation-based prompt generation method. During training, we evaluate the performance of the current policy on each reward branch, and these evaluation results are converted into text, which is then used as input for the prompt. Based on this prompt, the LLM selects the most appropriate weight adjustment rule from the rule repository to modify the weight calculation method for each branch. To ensure that the LLM understands the robot environment, task, and available rule options, we integrate the robot environment code, task description, and rule repository along with the policy evaluation content into a single prompt.

To address the limitations of human-designed rewards and enhance the training efficiency of our framework, we further introduce an auxiliary reward component. Drawing inspiration from the method used in Eureka \cite{ma2023eureka}, which employs large language models (LLMs) to design reward functions, we input the task description, environment code, original reward function code, and the optimization objectives of the framework as prompts to the LLM, which then designs an auxiliary reward component that promotes skill acquisition. This auxiliary reward is formulated before the start of training and incorporated into the multi-branch value network as one of the reward components.

Experimental results show that the AHRS method improves cumulative rewards across various tasks by approximately 6.48\% compared to Proximal Policy Optimization (PPO) baseline, and by around 5.52\% compared to the HD-PPO \cite{huang2022reward} method proposed in HDPG \cite{huang2022reward}.

\section{RELATED WORK}
\subsection{Hybrid Reward Reinforcement Learning}
To improve the efficiency of RL training and address the challenge of optimally approximating value functions in complex problems using low-dimensional representations, Van et al. proposed a method HRA (Hybrid Reward Architecture) \cite{van2017hybrid} that improves learning efficiency by decomposing the reward function. The design philosophy of HRA \cite{van2017hybrid} is derived from the Horde architecture \cite{sutton2011horde}, which allows multiple ``agents" (demons) to learn in parallel. The HRA \cite{van2017hybrid} method excels particularly in tasks with vast state spaces, such as Ms. Pac-Man, where it has achieved performance surpassing that of human players.

The Hybrid and Dynamic Policy Gradient (HDPG) method \cite{huang2022reward} builds upon the HRA \cite{van2017hybrid} framework and further introduces a dynamic weighting mechanism for hybrid policy gradients, which captures the dependencies between reward components and enhances learning efficiency. Unlike HRA \cite{van2017hybrid} , HDPG \cite{huang2022reward} is based on DDPG \cite{lillicrap2015continuous} and can handle tasks with continuous action spaces, thereby extending the applicability of HRA \cite{van2017hybrid} to different scenarios.

\subsection{Application of LLMs in Robot Learning}
The rapid advancement of large language models (LLMs) has led to significant breakthroughs in robotics, particularly in skill acquisition, reward function design, and task planning \cite{kim2024survey,wei2022chain,ahn2022can}. 
In the domain of robot skill acquisition, LLMs is applied to generate executable task plans and operational strategies through natural language \cite{guan2023leveraging,lin2023text2motion}, greatly enhancing the efficiency of skill acquisition. Approaches like ProgPrompt \cite{singh2023progprompt} and Code-As-Policies \cite{liang2023code} demonstrate LLMs' ability to generate task-specific code, enabling robots to generalize across diverse scenarios efficiently.

Recently, LLMs have been used to convert natural language instructions directly into reward functions, simplifying the design process for reinforcement learning tasks. Yu et al. generates parameterized reward functions via LLMs and leverage online optimization techniques to solve specific tasks \cite{yu2023language}. However, such reward functions may lack precision in low-level control tasks \cite{wang2023robogen}. To address this, EUREKA \cite{ma2023eureka} introduces interpretable white-box reward codes, improving the efficacy and transparency of reward functions in complex environments.
Beyond skill acquisition and reward function design, LLMs have also demonstrated strong capabilities in robot task planning. LLMs can generate high-level task plans \cite{brohan2022rt, kwon2024language} and dynamically adjust action plans based on feedback \cite{mandi2024roco}, enabling robots to adapt flexibly to tasks in open and complex environments \cite{liu2023llm+,sun2024interactive}. 

\begin{figure*}[t]
    \centering
    \includegraphics[width=0.80\textwidth, trim=85 130 85 65, clip]{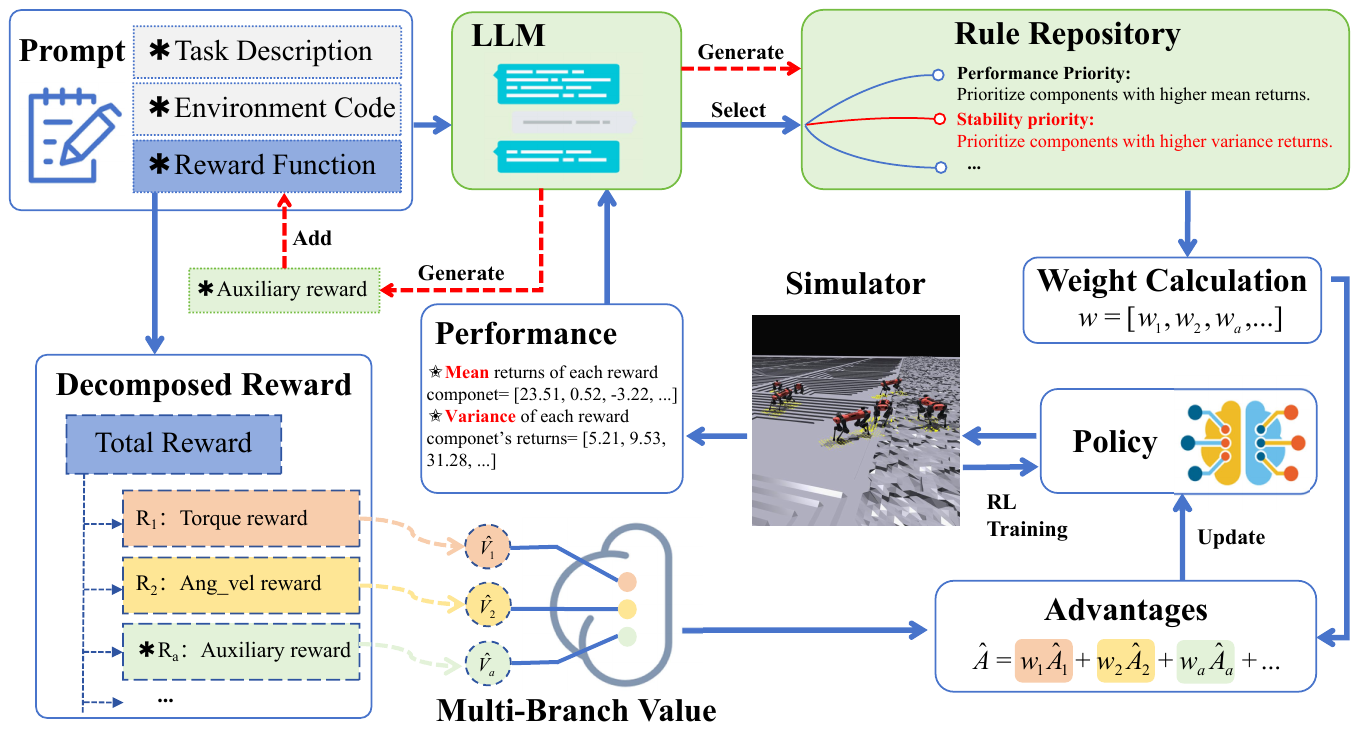}
    \caption{Overview of the proposed Automated Hybrid Reward Scheduling (AHRS) framework. It includes multi-branch value networks, the construction of dynamic weight rule repository, the selection of rules, and the generation of auxiliary reward functions.}
    \vspace{-10pt}
    \label{fig:framework}
\end{figure*} 

\section{BACKGROUND}

\subsection{Markov Decision Process(MDP)}

In reinforcement learning, the Markov Decision Process (MDP) models the interaction between an agent and its environment. At each time step \( t \), the agent observes the state \( s_t \in \mathcal{S} \), selects an action \( a_t \in \mathcal{A} \) according to policy \( \pi \), and receives a reward \( r \). The MDP is represented by \( (\mathcal{S}, \mathcal{A}, \mathcal{P}, \mathcal{R}, \gamma) \), where \( \mathcal{S} \) is the state space, \( \mathcal{A} \) is the action space, \( \mathcal{P} \) is the state transition function, \( \mathcal{R} \) is the reward function, and \( \gamma \) is the discount factor balancing immediate and future rewards.
The agent's goal is to learn a policy \( \pi_{\theta}(a_t \mid s_t) \) that maximizes the cumulative return by optimizing the policy gradient:
\begin{equation}
\label{eq:policy_gradient}
\nabla_{\theta} J(\pi_{\theta}) = \mathbb{E}_{\tau \sim \pi_{\theta}}\left[\sum_{t=0}^{T} A_{\pi}(s_t, a_t)\nabla_{\theta} \log \pi_{\theta}(a_t \mid s_t) \right]
\end{equation}
where \( A_{\pi}(s_t, a_t) = R_t - V(s_t) \) is the advantage estimation. Here, \( R_t \) is the estimate of cumulative return, calculated as: $R_t = \sum_{i=0}^{\infty} \gamma^{i} r_{t+i}$
and \( V(s_t) \) is the value estimate of state \( s_t \). In PPO \cite{schulman2017proximal}, the advantage function \( A_{\pi}(s_t, a_t) \) is estimated using Generalized Advantage Estimation (GAE) \cite{schulman2015high}. Due to its effectiveness across various tasks, PPO is selected as the baseline algorithm for this study.

\subsection{Hybrid Reward Architecture for Reinforcement Learning}

When reinforcement learning is applied to robotic skill learning tasks, multiple constraints are often involved, and the reward function typically consists of multiple components \cite{makoviychuk2021isaac}. Existing RL methods that sum the rewards to learn a single value function can limit the efficiency of policy optimization \cite{juozapaitis2019explainable, van2017hybrid}. To address this issue, van et al. proposed a Hybrid Reward Architecture \cite{van2017hybrid}, which decomposes the reward function $r$ into $K$ sub-reward components: $r_t = [r_{t,1}, r_{t,2}, ..., r_{t,K}]$, and learns a separate value function for each component. Learning such fine-grained value functions has been shown to effectively facilitate policy learning. HDPG \cite{huang2022reward} demonstrated this method is also applicable to the PPO algorithm \cite{schulman2017proximal}. Under this architecture, multiple advantage functions are learned, with each advantage function corresponding to a reward component: $A(s_t, a_t) = [A_{t,1}, A_{t,2}, \dots, A_{t,k}]$. The policy gradient is computed for each advantage function and then summed to obtain the final policy gradient:
\begin{equation}
\label{eq:HD_ppo_policy_gradient}
\nabla_{\theta} J(\pi_{\theta}) \approx \mathbb{E}_{s_t, a_t}\sum_{k=1}^{K} A_{\pi}^k(s_t, a_t) \nabla_{\theta} \log \pi_{\theta}(a_t \mid s_t)
\end{equation}

\section{METHOD}

The AHRS framework (Fig.~\ref{fig:framework}) integrates multi-branch value networks with LLMs for dynamic reward adjustment. Before RL training, a rule repository is generated with task-specific rules. The total reward is decomposed into sub-reward components (e.g., torque, angular velocity), each assigned to a value branch. During training, component performance metrics, such as the mean and variance of returns, are used to create prompts for the LLM, which selects the appropriate rule to compute dynamic weight coefficients. These weights optimize the policy, updated iteratively through RL. Additionally, an automatically generated auxiliary reward is introduced to further enhance learning. 

\subsection{Dynamic Policy Gradient}
Similar with HRA \cite{van2017hybrid}, we decompose the total reward $r_t$ of the environment and represent it as a vector $r_t$, with each reward component equipped with an independent value branch. 
In HDPG \cite{huang2022reward}, it is suggested that to capture potential dependencies between reward components, skills should be learned in a prioritized sequence. The motivation behind this approach is to encourage the agent to first master simpler components and then progressively learn more complex ones. Specifically, a set of weight coefficient vectors $w_t=[w_{t,1}, w_{t,2}, \dots, w_{t,K}]$ is assigned to each reward component according to a particular computation rule, then Eq. \ref{eq:HD_ppo_policy_gradient} can be rewritten as:
\begin{equation}
\label{eq:weighted_advantage_estimation}
\nabla_{\theta} J(\pi_{\theta})\approx \mathbb{E}_{s_t, a_t}\sum_{k=1}^{K}I_k\nabla_{\theta} \log \pi_{\theta}(a_t \mid s_t)
\end{equation}
where $I_k=w_k A_k$.
However, the design of such weight calculation rules relies heavily on human expertise. For different robots or tasks, specific rules must be crafted, which is clearly time-consuming and labor-intensive, making it difficult to meet the demands of real-world robotic applications. Intuitively, incorporating various rules could allow for more fine-grained weight adjustments, but the key challenge lies in how to construct a repository of diverse and effective rules that can address different environments and complex robotic tasks. Furthermore, determining which rules to apply in specific situations to enhance the efficiency of policy training is a central problem that we aim to investigate.

\subsection{Rule Repository Construction} \label{sec:Rule repository}
We propose a language-instructed rule generation method to construct a rule repository for dynamic weight calculation. Specifically, we input the robot's task description $T_t$, environment information $T_e$, and the code for each reward component $C_r$ in textual form into the prompt for the LLM. Additionally, to ensure that the LLM generates reasonable weight computation rules, we provide the rules proposed in the HDPG \cite{huang2022reward} method as examples, denoted as $E_\text{hdpg}$, for the LLM to reference, thereby enhancing the validity of the generated rules.
\begin{equation}
\label{eq:rule_generated_llm}
\mathcal{B}^n = \text{LLM}^g(T_t,T_e,C_r,E_\text{hdpg})
\end{equation}
where $\mathcal{B}^n$ represents the repository of rules generated by the LLM, with each rule expressed as a mathematical formulation for weight computation along with an explanation of the rule.
The detailed rule repository construction process and related rules are presented in Appendix Section A.
Leveraging the reasoning capabilities, LLM can construct a dynamic weight computation rule repository. This repository is built prior to the commencement of RL training.
The mathematical expression of the rule will be converted into text and code expression, which will be selected by the LLM in the subsequent training process.

\begin{algorithm}[t] 
\caption{AHRS: Automated Hybrid Reward Scheduling}
\label{alg:AHRS}
\begin{algorithmic}[1]
\REQUIRE Initial policy $\pi_0$, training epochs $N$, task description $T_t$, environment information $T_e$, reward code $C_r$.
\STATE \textcolor{lightgreen}{// Decompose total reward}
\STATE  $r_{total} = [r_1, r_2, ..., r_K]$
\STATE \textcolor{lightgreen}{// Generate rule repository and auxiliary reward}
\STATE  $\mathcal{B}^n = \text{LLM}^g(T_t,T_e,C_r,E_\text{hdpg})$
\STATE  $r_a = \text{LLM}^g(T_t,T_e,C_r)$
\FOR{$n = 1$ \textbf{to} $N$, \textbf{every} 100 \textbf{epochs}}
    \STATE Acquire policy performance $\textbf{S}^R_l$
    \STATE \textcolor{lightgreen}{// Select a rule from repository$\mathcal{B}^n$}
    \STATE $B_\text{selectsed} = \text{LLM}^s(T_t,T_e, C_r, \mathcal{B}^n, \textbf{S}^R_l, \textbf{S}^R_L)$
    \STATE Append $\textbf{S}^R_l$ to the queue $\textbf{S}^R_L$
    \STATE \textcolor{lightgreen}{// Generate a weight vector}
    \STATE $[w_1, \dots, w_a, \dots, w_K]=B_\text{selected}(\textbf{S}^R_l, \textbf{S}_\sigma)$
    \STATE Calculate advantages: $\sum_{k=1}^K (w_{t,k} A_{t,k})+w_{t,a}A_{t,a}$
\STATE Update policy $\pi$ through policy gradients

\ENDFOR

\STATE \textbf{Output:} Policy $\pi$
\end{algorithmic}
\end{algorithm}

\subsection{Automated Hybrid Reward Scheduling}
In HDPG \cite{huang2022reward}, the dynamic weights of reward components are calculated using fixed rules, which limits the flexibility and efficiency of policy optimization. By utilizing the rule repository, AHRS enables more flexible adjustments and switches between calculation rules during training, allowing for a more reasonable configuration of dynamic weights. 

\begin{table*}[t]
\centering
\begin{tabular}{ccccccc}
\toprule
Methods& \multicolumn{6}{c}{Task}\\
\cmidrule(lr{0.3em}){2-7}
 -& AnymalTerrain& Ant& ShadowHand& Quadcopter& AllegroHand&Cassie\\
\midrule
 PPO& 22.26$\pm$0.29& 9503.13$\pm$411.42& 7000.35$\pm$311.44& 1267.86$\pm$6.48& 4501.63$\pm$130.31& 1.55$\pm$0.08\\
 HD-PPO& 23.25$\pm$0.18& 8852.27$\pm$501.61& 7349.86$\pm$50.80& 1291.19$\pm$25.25& 4409.02$\pm$57.03& 1.61$\pm$0.08\\
 AHRS w/o A& 23.47$\pm$0.23& 9561.81$\pm$305.92& 7448.22$\pm$14.79& 1296.73$\pm$65.79& 4582.46$\pm$216.52&\textbf{1.69$\pm$0.08}\\
 AHRS& \textbf{23.64$\pm$0.07}& \textbf{10118.23$\pm$318.22}& \textbf{7642.38$\pm$91.73}& \textbf{1344.10$\pm$11.95}& \textbf{4646.92$\pm$103.22}& 1.67$\pm$0.01\\
\bottomrule
\end{tabular}
\caption{Accumulative reward comparison across six tasks, presented as mean $\pm$ standard deviation of returns. Our method (AHRS w/o A, AHRS) consistently achieves superior performance across all tasks, outperforming other methods. \textbf{Bolded} numbers indicate the best performance.}
\label{tab:maintable}
\vspace{-15pt}
\end{table*}

Skill prioritization, or adjusting the importance of reward components, is key to effective policy optimization. LLMs analyze the performance of different skills and their impact on overall policy, identifying which skills are most critical at each stage. This enables timely adjustments to skill weights, enhancing learning efficiency and policy effectiveness. The proposed framework evaluates the policy's performance on each reward component during training, feeding this data to the LLM, which then selects appropriate weight computation rules to meet the needs of policy optimization.

\begin{figure}[t]
    \centering
    \includegraphics[width=0.9\columnwidth, trim=85 140 85 120, clip]{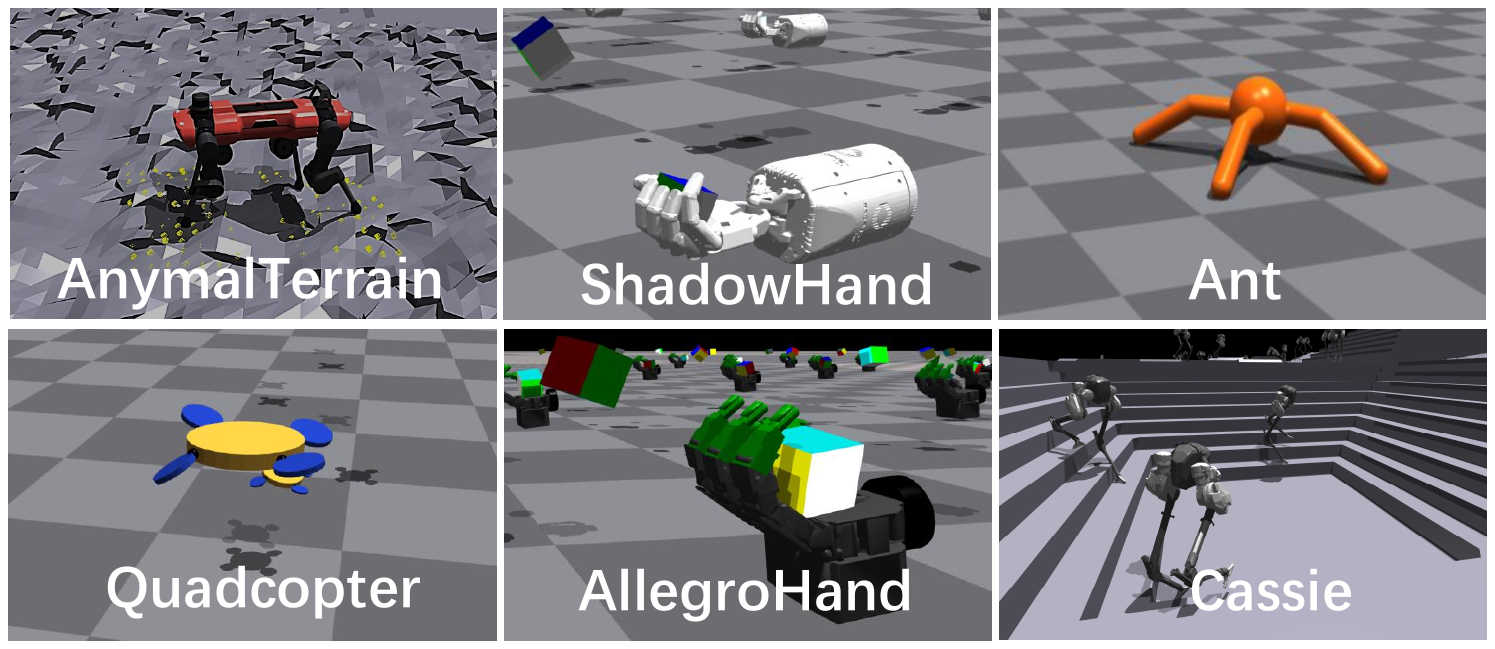}
    \caption{Illustrations of the six tasks in this experiment: Ant, ShadowHand, AnymalTerrain, AllegroHand, 
    Quadcopter, Cassie}
    \label{fig:task}
    \vspace{-15pt}
\end{figure} 

In practical implementation, task description $T_t$, environment description $T_e$, reward component information $C_r$, and the rule set $\mathcal{B}^n$ for weight computation, as constructed in Sec. \ref{sec:Rule repository}, are used as prompt information. Additionally, the current policy's performance on each reward component, $\textbf{S}^R_t$, and the historical performance of policy, $\textbf{S}^R_T$, are provided to the LLM for reasoning. The LLM analyzes the current policy's performance and, based on this information, selects the most appropriate weight computation rule from $\mathcal{B}^n$. The method for generating the weights is expressed as follows:
\begin{equation}
\label{eq:weight_computation_llm}
B_\text{selected} = \text{LLM}^s(T_t,T_e, C_r, \mathcal{B}^n, \textbf{S}^R_l, \textbf{S}^R_L)
\end{equation}
Here, $\textbf{S}^R_l = [R_l^1, R_l^2, \dots, R_l^K]$ represents the average estimate return of the current policy across each reward component, and $\textbf{S}^R_L$ denotes the collection of historical returns during the training process: $\textbf{S}^R_L = [\textbf{S}^R_{l-1}, \textbf{S}^R_{l-2}, \dots, \textbf{S}^R_{l-L}]$, where $L$ is the length of the historical data.
Based on the rule $B_\text{selected}$ obtained from Eq. \ref{eq:weight_computation_llm}, the priority weights for the policy gradient can be further calculated.
\begin{equation}
\label{eq:weight_calculation}
[w_1, \dots, w_k, \dots, w_K]=B_\text{selected}(\textbf{S}^R_l, \textbf{S}_\sigma)
\end{equation}
where $\textbf{S}_\sigma = [\sigma^1, \sigma^2, \dots, \sigma^K]$ represents the variance of each return branch in $\textbf{S}^R_l$.
The specific prompt for LLM rule selection has been placed in Appendix Section A.

\subsection{Auxiliary Reward}

In complex and high-dimensional robotic tasks, human-designed reward functions may not always enable efficient policy training. To enhance the efficiency of skill learning in robots, we introduce an auxiliary reward component. The aim is to leverage the LLM's understanding of the training task and optimization objectives to generate auxiliary reward components that enhance the reward signals during training: $r_a = \text{LLM}^a(T_t, T_e, C_r)$.
The detailed prompt for generating auxiliary rewards has been provided in Appendix Section A.
The approach of using LLMs to design reward function has been proven feasible in Eureka \cite{ma2023eureka}. This auxiliary reward, added as a new component, is incorporated into the multi-branch value network for training. As a result, the decomposed reward function vector for each task can be reformulated as $r_t = [r_{t,1}, r_{t,2}, \dots, r_{t,K}, r_{t,a}]$. Similarly, the weight coefficient vector for each reward component can be expressed as $w_t = [w_{t,1}, w_{t,2}, \dots, w_{t,K}, w_{t,a}]$. Therefore, Eq. \ref{eq:weighted_advantage_estimation} can be reformulated as follows:
\begin{equation}
\label{eq:final_advantage}
\nabla_{\theta} J(\pi_{\theta}) \approx \mathbb{E}_{s_t, a_t}(I_a+\sum_{k=1}^{K}I_k)\nabla_{\theta} \log \pi_{\theta}(a_t \mid s_t)
\end{equation}
Here, $I_a = w_a A_a$ represents the weighted advantage estimation for the auxiliary reward component. In practice, the LLM directly generates the code for the auxiliary reward. The overall implementation of AHRS is shown in Alg. \ref{alg:AHRS}.

\section{EXPERIMENTS}

\begin{figure*}[t]
    \centering
    \includegraphics[width=0.80\textwidth, trim=150 180 140 150, clip]{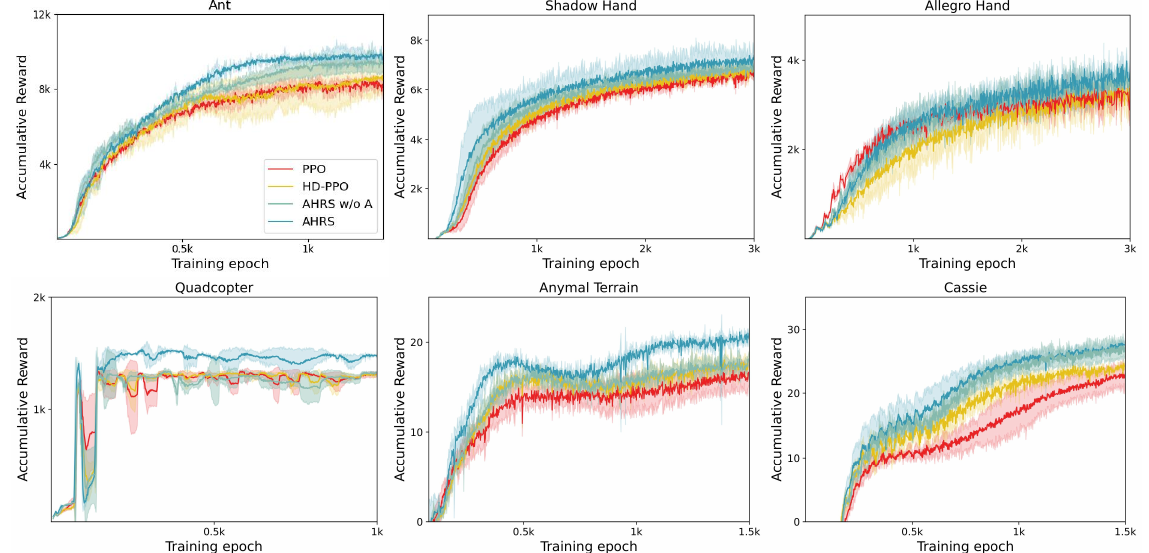}
    \caption{The learning curves show the variation in accumulative rewards across multiple environments (Ant, ShadowHand, AllegroHand, Quadcopter, AnymalTerrain, Cassie) over training epochs. Different colored lines represent different methods: PPO (red), HD-PPO  (yellow), AHRS w/o A (cyan), and AHRS(blue). The solid lines represent the mean values for each method, while the shaded areas indicate the standard deviation. }
    \label{fig:train}
\end{figure*}

\subsection{Experimental Setting}

All experimental tasks in this paper are conducted within the Isaac Gym environment \cite{makoviychuk2021isaac}, as shown in Fig. \ref{fig:task}, using the Proximal Policy Optimization (PPO) algorithm to train multiple robotic tasks. 
The initial reward settings and parameters follow the default configurations provided by Isaac Gym \cite{makoviychuk2021isaac}. The selected task scenarios encompass a wide range of robotic operations, including multi-legged robots (Ant, AnymalTerrain, Cassie), robotic arms (ShadowHand, AllegroHand), and drone control tasks (Quadcopter). These tasks cover various types of robot control problems, aiming to validate the effectiveness of the proposed algorithm across multiple complex tasks.

The methods involved in the experiments conducted in this paper are as follows. PPO and HD-PPO \cite{huang2022reward} are the two baseline methods, while AHRS is the proposed approach. The specific descriptions are as follows:

\textbf{PPO:} The standard PPO algorithm, which optimizes the policy by using a summed reward function.

\textbf{HD-PPO:} Unlike traditional PPO, HD-PPO \cite{huang2022reward} employs a multi-branch value network with dynamic weights to adjust the priority of each branch, where the weight calculation rules are manually designed.

\textbf{AHRS:} The proposed method.

\textbf{AHRS w/o A:} The proposed method without auxiliary reward components is designed to verify the effectiveness of the auxiliary rewards.

\begin{table*}[t]
\centering
\begin{tabular}{ccccccc}
\toprule
Methods& \multicolumn{6}{c}{Task}\\
\cmidrule(lr{0.3em}){2-7}
 -& AnymalTerrain& Ant& ShadowHand& Quadcopter& AllegroHand&Cassie\\
\midrule
 AHRS-R& 22.61$\pm$0.41& 9301.81$\pm$589.96& 6817.18$\pm$644.55& 1300.45$\pm$35.35& 4017.80$\pm$786.03& 1.58$\pm$0.07\\
 AHRS-D& 23.01$\pm$0.25& 9281.33$\pm$325.58& 7099.66$\pm$195.29& 1220.33$\pm$31.00 & 3847.26$\pm$353.29& 1.62$\pm$0.04\\
 AHRS& \textbf{23.64$\pm$0.07}& \textbf{10118.23$\pm$318.22}& \textbf{7642.38$\pm$91.73}& \textbf{1344.10$\pm$11.95}& \textbf{4646.92$\pm$103.22}& \textbf{1.67$\pm$0.01}\\
\bottomrule
\end{tabular}
\caption{In the ablation study, the AHRS-R experiment involves randomly selecting a rule from the set of rules used in AHRS for each rule adjustment. In contrast, the AHRS-D experiment directly assigns weights based on the task description and branch performance as provided by the LLM, bypassing the rule-based calculation process. Bolded numbers indicate the best performance.}
\label{tab:ablationtable}
\vspace{-15pt}
\end{table*}

\begin{figure}[t]
    \centering
    \includegraphics[width=0.95\columnwidth, trim=170 180 170 180, clip]{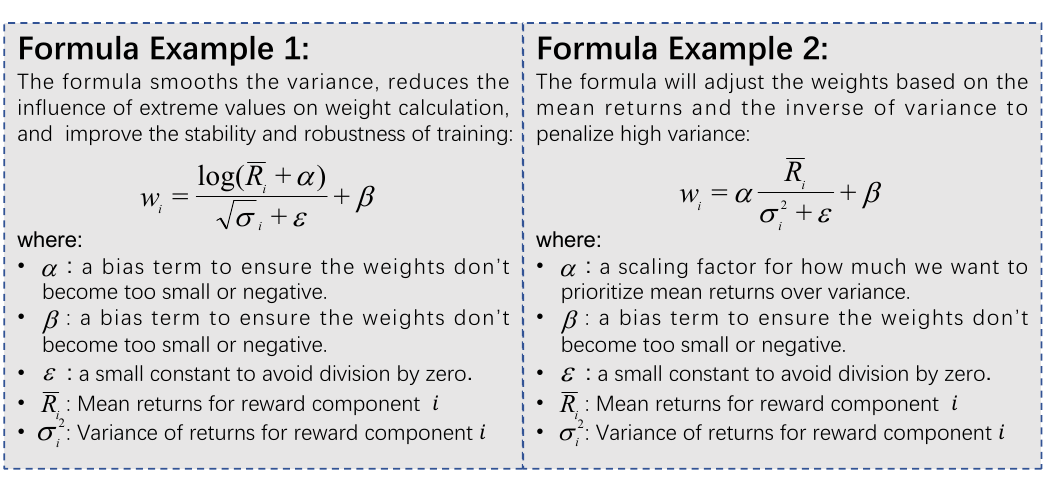}
    \caption{Examples of LLM-generated rules include: Formula 1, which uses logarithmic means and adjusted variances to smooth out extreme values, and Formula 2, which balances average return and variance, scaled by $\alpha$ and offset by $\beta$, to prioritize stable, high-performing components.}
    \label{fig:Rule generated form LLM}
\end{figure}

AHRS employs multiple pre-generated weight calculation rules during training. The LLM (GPT-4o in this work) provides the mathematical formulations and characteristics of each rule, aiding the optimization of multi-branch networks, as shown in Fig. \ref{fig:Rule generated form LLM}. Every 100 epochs, the LLM selects the optimal rule based on the training data for each reward component, determining the weights for each reward branch and enabling adaptive adjustment of the reward structure.

\textbf{Parameter setting.}
We set a base weight of $w_{\text{base}} = 0.5$ for each reward component. Thus, in the HD-PPO \cite{huang2022reward}, AHRS w/o A, and AHRS experiments, the final weight for each component is $w^k = w_{\text{base}} + w_{\text{calculated}}^k$, where $w_{\text{calculated}}^k$ is the value computed by the corresponding weight calculation rule. The maximum training iterations for each task follow the settings of IsaacGym \cite{makoviychuk2021isaac}: 1500 epochs for Ant, AnymalTerrain, and Cassie; 1000 epochs for Quadcopter; and 3000 epochs for ShadowHand and AllegroHand. The historical data length $L$ is 5 in AHRS and AHRS w/o A.

\subsection{Comparison with Baseline Methods}
We present the results of six robotic tasks in Tab. \ref{tab:maintable}. The proposed AHRS w/o A and AHRS methods show significant improvements over PPO and HD-PPO. For example, AHRS improves by 6.01\% over PPO in the Quadcopter task, 3.23\% in AllegroHand, 9.0\% in ShadowHand, and 7.74\% in Cassie. The results clearly indicate that AHRS achieves optimal performance in most tasks.

Tab. \ref{tab:maintable} also shows that HD-PPO underperforms in the Ant and AllegroHand tasks, with scores 7.8\% and 3\% lower than PPO. While HD-PPO benefits from rule-based structures in some cases, it lacks adaptability in variable environments. Comparisons reveal that AHRS w/o A consistently outperforms HD-PPO, e.g., achieving a cumulative reward of 448.22$\pm$14.79 in ShadowHand, compared to HD-PPO's 7349.86$\pm$50.80 and PPO's 7000.35$\pm$311.44. This demonstrates that dynamic adjustments like AHRS outperform fixed rule designs, especially in complex environments. The flexibility of AHRS w/o A allows for better performance.

In AHRS, the LLM generates auxiliary reward components tailored to the task, enhancing learning efficiency. For instance, AHRS shows a 4.5\% improvement over AHRS w/o A in the Quadcopter task and 5.82\% in the Ant task, proving the method’s effectiveness.

Fig. \ref{fig:train} shows the training performance of each algorithm. AHRS w/o A and AHRS converge faster in the early stages of most tasks. In the Ant task, AHRS w/o A surpasses HD-PPO and PPO within the first 500 iterations and continues to improve, demonstrating high learning efficiency. AHRS also exhibits lower training variance in the Ant and AnymalTerrain tasks, indicating greater stability. In the AllegroHand task, AHRS w/o A and AHRS improve at a similar rate to PPO in early and mid-stages but surpass PPO in later stages, showing that LLM rule adjustments help models adapt better in later phases.

\subsection{Ablation studies}
In this section, we examine two key points: First, we compare LLM's dynamic rule adjustment with random rule selection (AHRS-R) to evaluate its impact on performance. Second, we compare LLM-based rule selection with direct weight generation (AHRS-D) to assess the necessity of constructing a rule set.

\textbf{Random Rule.}
AHRS-R selects weight calculation rules randomly during training without relying on fixed or intelligent selection methods. This serves as a control to evaluate the effectiveness of LLM's dynamic adjustment. If random selection performs similarly to or better than LLM or fixed rules, it would suggest that intelligent rule selection provides little benefit. As shown in Tab. \ref{tab:ablationtable}, random selection generally performs worse than LLM-chosen or fixed rules. For instance, in the AnymalTerrain task, random selection achieves a cumulative reward of 22.61$\pm$0.41, lower than AHRS's 23.64$\pm$0.07. In the ShadowHand task, AHRS achieves 7642.38$\pm$91.73, about 12.11\% higher than random selection. Overall, random selection underperforms compared to LLM dynamic selection, highlighting LLM's positive impact.

\textbf{Weight generation from LLM directly.}
AHRS-D examines LLM's ability to directly generate weights from feedback. In this setup, the LLM infers weights based on task, environment, and historical performance, without using any predefined rules. As shown in Tab. \ref{tab:ablationtable}, rule-based weight generation (particularly with LLM's dynamic adjustment) generally outperforms direct LLM weight generation in most tasks. While LLM excels at reasoning from feedback, directly generating weights leads to instability and higher variance in complex tasks. In contrast, AHRS dynamically adjusts rules throughout training, enhancing performance, while direct weight generation lacks this adaptive strategy, resulting in weaker performance.

\section{CONCLUSION}
In this work, we propose the Automated Hybrid Reward Scheduling (AHRS) framework to address the inefficiencies of traditional reinforcement learning methods in high-degree-of-freedom robotic tasks. By dynamically adjusting the learning intensity of each reward component through the use of Large Language Models (LLMs), the AHRS framework facilitates a more structured and efficient skill acquisition process. The integration of a multi-branch value network, guided by LLM-generated rules for weight adjustment of each reward component, enables more effective policy optimization. Experimental results confirm the effectiveness of this approach, showing an average performance improvement of 6.48\% across various complex robotic tasks, highlighting the potential of AHRS to enhance the learning capabilities of high-degree-of-freedom robots.
Our current task focuses on validating our method in simulation. Future research will involve sim-to-real experiments to assess its feasibility and safety in real-world scenarios.

\bibliographystyle{IEEEtran}
\bibliography{ref}

\clearpage

\appendix
This appendix provides a detailed explanation of the prompt strategy design used in AHRS training in Section A. In Section B, additional ablation studies on the auxiliary reward mechanism are conducted to verify the effectiveness of the dynamic scheduling mechanism. Section C expands the scope of experiments by introducing more complex task environments to further validate the training efficiency advantages of AHRS. 
\subsection{Prompts Detail}
\textbf{Prompt for building a rule repository:} Prior to training, we asked the LLM for the weight calculation rules used during training through the following prompt, which does not change as the task environment changes.

\begin{mdframed}[backgroundcolor=lightgreen!20, linecolor=lightgray, leftmargin=0cm, rightmargin=0cm, innerleftmargin=0.3cm, innerrightmargin=0.3cm]

\color{black}
{\textbf{Build rule repository}}\\
I have an optimization problem in reinforcement learning (RL) that I'd like you to solve. Please use your advanced problem-solving and analytical skills to provide a thorough and accurate solution.
Here is the optimization problem you need to solve:\\
====================================\\
\textbf{Task description:} You are an expert in reinforcement learning algorithms and need to figure out how to adjust the weight coefficients of different reward components during training to enhance performance.The task involves training a robot to complete the goal using n reward components. \\
====================================\\
\textbf{Algorithm framework:}The algorithm framework builds on Proximal Policy Optimization (PPO) but decomposes the total reward into n components.

...

Think step and step, your goal is to write at least six useful weight generation rules (mathematical representation) to generate weight coefficients that will help the agent learn the task described in text. And then, you need to write it as python code.\\
====================================\\
When you write your weight generation rule, assuming that your inputs are mean returns and var returns, which are come from each reward component.

I will give an example and you can refer to this example to provide a better weight generation rule: 

(HDPG's Rule)

...

\textbf{Some tips may be helpful for you:} You can consider adding simple hyper-parameter, or tweaking the normalization method.

\end{mdframed}

\textbf{Rule repository of AHRS:} The detailed rules we use in AHRS and their mathematical expression are shown in Figure 5.

\begin{figure}[t]
    \centering
    \includegraphics[scale=0.5, trim=50 170 50 40, clip]{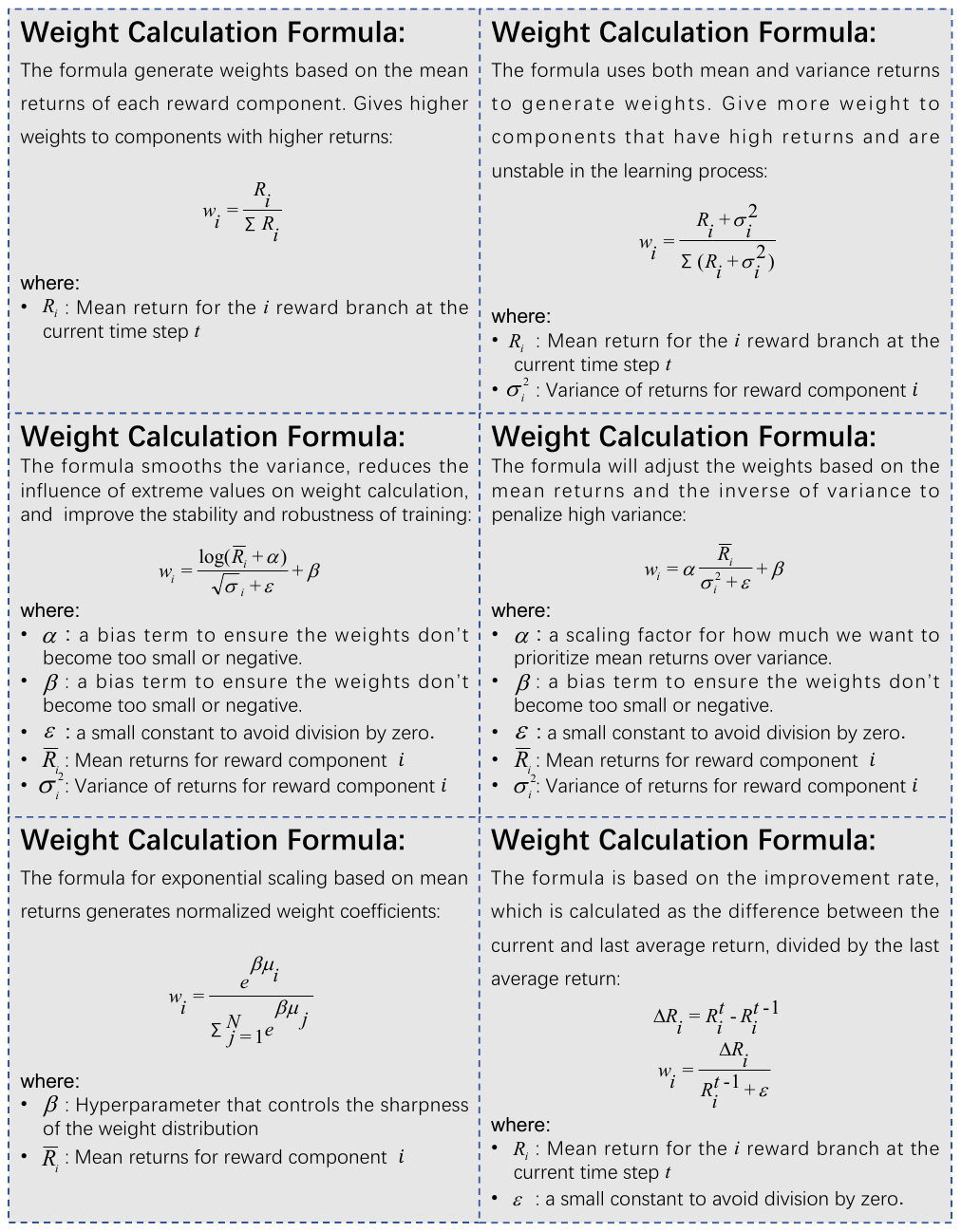}
    \caption{Rule repository of AHRS.}
    \label{fig:Rule repositoty}
\end{figure}

\textbf{Prompt for Auxiliary reward:} Before the start of each task, we will put the task's environment description and environment code, as well as the initial reward function in the task into the prompt to ask the LLM, in order to get an auxiliary reward that can improve the learning effect. Take the Quadcopter task as an example: \\

\begin{mdframed}[backgroundcolor=yellow!20, linecolor=lightgray, leftmargin=0cm, rightmargin=0cm, innerleftmargin=0.3cm, innerrightmargin=0.3cm]

\color{black}
\textbf{Add Auxiliary reward}\\
You are an expert in reinforcement learning algorithms. You should help me write proper auxiliary reward functions to train a Quadcopter robot  agent with reinforcement learning to complete the described task.\\
====================================\\
\textbf{Task description and Reward function:} The goal of the Quadcopter task is to navigate efficiently to a target while maintaining stable flight and minimizing excessive spinning.Here are the reward functions for the quadcopter task: 
\color{deepblue}
\begin{verbatim}
@torch.jit.script
def compute_quadcopter_reward(
root_positions, root_quats, 
root_linvels, root_angvels, reset_buf,
progress_buf, max_episode_length):
...
\end{verbatim}
\color{black}
====================================\\
\textbf{Explaination for each reward components of the reward function:}

1.pos reward: Rewards the quadcopter based on its distance to a target position, encouraging it to get closer to the target. The reward increases as the distance decreases.

2.pos up reward: Combines the position reward with the uprightness reward, promoting not only reaching the target but also maintaining an upright orientation.

3.pos pinnage reward: Combines the position reward with the spinning reward, encouraging the quadcopter to get closer to the target while minimizing excessive spinning.\\
====================================\\
\textbf{Reward function requirements:}\\
You should write an auxiliary reward function based on the reward function I gave to help the agent perform its task better.

The auxiliary reward function you add should not change the reward component of the original reward function, but rather add on top of it.\\
\textbf{Output Requirements: }

1.The reward function should be written in Python 3.7.16. 

2.Output the code block only. \textbf{Do not output anything else outside the code block}. 

3.You should include \textbf{sufficient comments} in your reward function to explain your thoughts, the objective and \textbf{implementation details}. The implementation can be specified to a specific line of code. 

4.If you need to import packages (e.g. math, numpy) or define helper functions, define them at the beginning of the function. Do not use unimported packages and undefined functions.

Output format Strictly follow the following format. \textbf{Do not output anything else outside the code block}: 

\color{deepblue}
\begin{verbatim}
Def compute_reward(self):
# Thoughts: 
# (...) 
# (import packages and define helper 
functions) 
    import numpy as np ... 
 ... (reward function)
\end{verbatim}
\color{black}
====================================\\
Now write a auxiliary reward functions based on the reward function I have given. Then in each iteration, I will use the reward function to train an RL agent, and test it in the environment. 

I will give you possible reasons of the failure found during the testing, and you should modify the reward function accordingly.
\end{mdframed}

\begin{figure*}[t]
    \centering
    \includegraphics[width=0.75\textwidth, trim=50 145 80 130, clip]{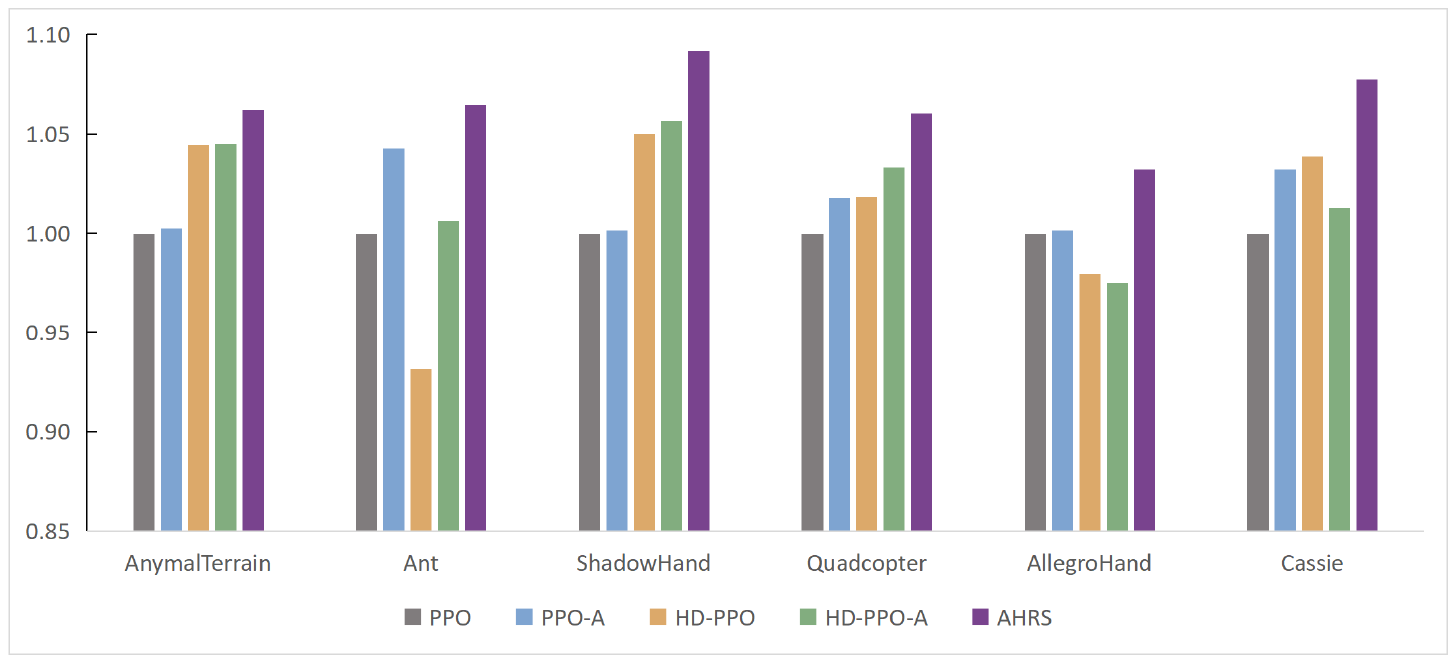}
    \caption{This figure compares the performance of five reinforcement learning algorithms—PPO, PPO-A, HD-PPO, HD-PPO-A, and AHRS—across six tasks: AnymalTerrain, Ant, ShadowHand, Quadcopter, AllegroHand, and Cassie. The bar chart illustrates the average performance differences of each algorithm relative to PPO, which serves as the baseline (represented by the horizontal dashed line).}
    \label{fig:train}
\end{figure*}

\textbf{Prompt for rules' selection:} During the training process, we summarize the performance of the agent (the returns of each reward component) every 100 epochs, feed back to the LLM, and then the LLM adjusts the weight calculation rules.The rule repository used in AHRS is also included in this prompt. 

\begin{mdframed}[backgroundcolor=lightgray!20, linecolor=lightgray, leftmargin=0cm, rightmargin=0cm, innerleftmargin=0.3cm, innerrightmargin=0.3cm]
\color{black}
\textbf{Rules' selection}\\
I have an optimization problem in reinforcement learning (RL) that I'd like you to solve. Please use your advanced problem-solving and analytical skills to provide a thorough and accurate solution.\\
Here is the optimization problem you need to solve:\\
====================================\\
\textbf{Task description:}  The task involves guiding ...

Their physical meanings are as follows:

1.pos reward: This reward ...

2.pos up reward: This reward ...

3.pos pinnage reward:This component ...

...

During different training stages, the importance of each reward component varies. Therefore, learning to adjust the weights of these components at different stages is crucial for more efficient policy training.\\
====================================\\
\textbf{Algorithm framework:} The algorithm framework is ...\\
====================================\\
\textbf{Environment description:}
\color{deepblue}
\begin{verbatim}
The Python environment is class 
Quadcopter(VecTask): 
def compute_observations(self):
    ...
\end{verbatim}
\color{black}

I'll provide arrays representing the mean returns, variance returns and weights obtained by each reward component in both the current and historical epochs.\\
====================================\\
\textbf{Goal:} \\
Please refer to the provided mean returns, variance returns, and weights for each reward component. Think step-by-step and consider the importance of each reward in improving the policy. Based on these datas, determine the best rule for generating weights that will benefit the current training epoch. You need to take into account the advantages and disadvantages of these rules.\\
====================================\\
\color{brown}
\textbf{Proposed rules:}\\
\textbf{1.Mean Returns Only:}\\
Generate weights based on the mean returns of each reward component.\\
Prioritize components with higher mean returns.\\
\textbf{2.Variance Returns Only:}\\
Generate weights based on the variance returns of each reward component.\\
Prioritize components with higher variance returns.\\
\textbf{3.Combined Mean and Variance Returns:}\\
Use both mean and variance returns to generate weights:
\begin{verbatim}
weight = mean_returns + var_returns
\end{verbatim}
\textbf{4.Improvement Rate Only:} ...\\
...\\
\color{black}
You need to choose the best method from the given options and tell me its serial number.\\
====================================\\
\textbf{Additional Context:}\\
You provided a suggestion for generating weights 100 epochs ago, and I used this suggestion to generate the weight coefficients. Consider this historical information when choosing the rules to generate weights for the current training stage.\\
====================================\\
\textbf{Output Format:}\\
Use the tilde symbol (~) at the beginning and end of the output serial number.
**Ensure the output is an integer and one of 1, 2, 3, 4, 5, 6, 7 or 8.**
Example: ~[1]~\\
====================================\\
\textbf{Current Data:}\\
Mean Returns: [...]
Variance Returns: [...]\\
====================================\\
Historical Mean Returns and Variance Returns (every 100 epochs) : [...]
\end{mdframed}

\begin{table}
\centering
\begin{tabular}{ccc}
\toprule
Methods& \multicolumn{2}{c}{Task}\\
\cmidrule(lr{0.3em}){2-3}
 -& ShadowHandScissors&ShadowHandBottleCap\\
\midrule
 PPO& 184.02$\pm$23.17& 238.88$\pm$17.85\\
 HD-PPO& 186.46$\pm$23.51& 234.31$\pm$31.23\\
 AHRS& \textbf{207.39$\pm$8.69}& \textbf{253.76$\pm$10.65}\\
\bottomrule
\end{tabular}
\caption{Accumulative reward comparison across two tasks, presented as mean $\pm$ standard deviation of returns. AHRS also maintains performance gains on more complex tasks. \textbf{Bolded} numbers indicate the best performance.}
\label{tab:maintable}
\vspace{-15pt}
\end{table}

\subsection{Ablation experiments on auxiliary rewards}

In AHRS, auxiliary rewards are also an important contribution. To further investigate their impact, we conducted ablation experiments on auxiliary rewards. The specific experimental settings are as follows:

\textbf{PPO-A}. We directly added auxiliary rewards to the initial static rewards and trained the agent using PPO, aiming to verify whether auxiliary rewards play a dominant role in improving training efficiency in the AHRS method.

\textbf{HD-PPO-A}. We introduced auxiliary rewards, decomposed the rewards, and used HD-PPO with a fixed weight calculation rule to train the agent, in order to validate the effectiveness of dynamic scheduling. The experimental results are as follows:

As shown in Fig. 6, PPO-A achieved an average performance improvement of approximately 2.34\%compared to PPO, while our method, AHRS, achieved an average improvement of approximately 6.48\% over PPO. Furthermore, AHRS, which incorporates a dynamic scheduling mechanism, outperformed HD-PPO-A (fixed weight calculation rule) by an average of approximately 4.21\%.

Overall, directly adding LLM-generated auxiliary rewards to static rewards does not significantly improve training efficiency. Additionally, the comparison between AHRS and HD-PPO-A effectively demonstrates that, after excluding the impact of auxiliary rewards, the effectiveness of dynamic scheduling remains evident.The experimental results of PPO-A and HD-PPO-A show that the value of auxiliary reward in AHRS can only be brought into full play with the combination of reward decomposition and dynamic scheduling strategy.

\subsection{Additional Experiment}

Building upon the original tasks, we further introduced two more complex tasks, ShadowHandScissors\cite{chen2023bi} and ShadowHandBottleCap\cite{chen2023bi}, to evaluate the effectiveness of AHRS in improving training efficiency. ShadowHandScissors requires both hands to cooperate to open the scissors. ShadowHandBottleCap involves two hands and a bottle and requires to hold the bottle with one hand and open the bottle cap with the other hand.

As shown in Table III, AHRS achieves an average performance improvement of approximately 9.45\% over PPO and 9.76\% over HD-PPO in these two tasks. The experimental results demonstrate that even in more complex environments, AHRS can still enhance training efficiency, verifying the generalization capability of the method.

\end{document}